\documentclass[letterpaper]{article} 
\usepackage{aaai24}  
\usepackage{times}  
\usepackage{helvet}  
\usepackage{courier}  
\usepackage[hyphens]{url}  
\usepackage{graphicx} 
\usepackage{natbib}  
\usepackage{caption} 
\usepackage{algorithm}
\usepackage{algorithmic}

\usepackage{newfloat}
\usepackage{listings}
\DeclareCaptionStyle{ruled}{labelfont=normalfont,labelsep=colon,strut=off} 
\floatstyle{ruled}
\newfloat{listing}{tb}{lst}{}
\floatname{listing}{Listing}
\title{Trust the Process: Zero-Knowledge Machine Learning to Enhance Trust in Generative AI Interactions}
\author {
    Bianca-Mihaela Ganescu\textsuperscript{\rm 1},
    Jonathan Passerat-Palmbach\textsuperscript{\rm 1, 2}
}
\affiliations {
    \textsuperscript{\rm 1}Imperial College London\\
    \textsuperscript{\rm 2}Flashbots\\
    bianca-mihaela.ganescu19@imperial.ac.uk, j.passerat-palmbach@imperial.ac.uk
}

\usepackage[utf8]{inputenc} 
\usepackage[T1]{fontenc}    
\usepackage{url}            
\usepackage{booktabs}       
\usepackage{amsfonts}       
\usepackage{nicefrac}       
\usepackage{microtype}      
\usepackage{xcolor}         

\begin{document}

\maketitle

\begin{abstract}
Generative AI, exemplified by models like transformers, has opened up new possibilities in various domains but also raised concerns about fairness, transparency and reliability, especially in fields like medicine and law. This paper emphasizes the urgency of ensuring fairness and quality in these domains through generative AI. It explores using cryptographic techniques, particularly Zero-Knowledge Proofs (ZKPs), to address concerns regarding performance fairness and accuracy while protecting model privacy. Applying ZKPs to Machine Learning models, known as ZKML (Zero-Knowledge Machine Learning), enables independent validation of AI-generated content without revealing sensitive model information, promoting transparency and trust. ZKML enhances AI fairness by providing cryptographic audit trails for model predictions and ensuring uniform performance across users. We introduce snarkGPT, a practical ZKML implementation for transformers, to empower users to verify output accuracy and quality while preserving model privacy. We present a series of empirical results studying snarkGPT's scalability and performance to assess the feasibility and challenges of adopting a ZKML-powered approach to capture quality and performance fairness problems in generative AI models.
\end{abstract}

\section{Introduction}\label{introduction}

The latest developments in Generative AI have opened new possibilities
for AI to become more pervasive in multiple fields. For example, GPT-4
can now pass internationally recognized exams in medicine and law
\cite{gpt4Medical, gpt4}. Such advancements in transformer-based models
introduce new challenges, particularly in ensuring the uniform quality
and execution of these models for all users when deployed and accessed
through remote cloud APIs. This notion is commonly known as performance
fairness \cite{Jiang_Roth_Li_Yang_Zhao_Nath_Xu_Dou_Xu_2023}.

Over the past few years, the field of AI fairness has witnessed
substantial progress in addressing bias mitigation within machine
learning models. This collective effort has yielded methods that strive
to create fairer models, reducing the impact of biases in various
applications. This paper
explores a novel algorithmic approach that complements AI fairness
strategies by focusing on ensuring the uniform performance and execution of deployed AI models across users.

We present a protocol that empowers users to
engage with remote servers hosting AI models, ensuring that the model
they interact with behind a black-box API is indeed the precise model
they have requested from the provider.

The implications of this approach extend across domains, with particular
relevance in fields such as healthcare and finance, where the outcomes
produced by AI models can profoundly affect the well-being of end-users,
especially those belonging to underrepresented populations.

Our approach leverages zkSNARKs (Zero-Knowledge Succinct Non-interactive
ARgument of Knowledge), a cryptographic tool that bolsters the integrity
of computations while preserving user privacy by not disclosing
sensitive information during the verification process.

Recent developments have given rise to a burgeoning body of research
known as ZKML (Zero-Knowledge Machine Learning)
\cite{feng2021zen,lee2020vcnn, https://doi.org/10.48550/arxiv.2210.08674}, which focuses on leveraging zkSNARKs to
verify the integrity of the inference pass in AI models. ZKML serves two
primary purposes:

\begin{enumerate}
\def\labelenumi{\arabic{enumi}.}
\item
  \textbf{Verification of Model Output:} In scenarios where users submit
  their data to remote services hosting AI models, ZKML provides users
  with a cryptographic proof that the output they receive originates
  from the exact model they have requested or paid for. This is
  particularly pertinent in Generative AI applications like ChatGPT,
  where multiple subscription tiers promise varying levels of model
  performance. Currently, users lack guarantees that the output aligns
  with their subscription tier.
\item
  \textbf{Enhancing Data Privacy:} ZKML is also valuable for
  enhancing data privacy. In cases where users are unwilling or unable
  to upload their data to remote third-party models, model owners can
  deploy models directly to users' devices. While this deployment
  strategy is unproblematic in scenarios like predictive text on a
  smartphone keyboard, more sensitive situations exist, such as
  financial services' Know Your Customer (KYC) processes. ZKML enables
  model providers to confidently deploy their models to users' devices,
  ensuring that the model's predictions are genuinely derived from their
  model and not simulated by the user using alternative software. The
  combination of data locality and the Zero-Knowledge property preserves
  user data privacy.
\end{enumerate}

In the context of AI fairness, ZKML directly contributes to performance
fairness, ensuring that all users of a particular service experience consistent quality. Model owners need not reveal their model's weights,
preserving their valuable intellectual property. Instead, they can
publish a model fingerprint that is verified as an additional step
during user output generation. This has far-reaching implications,
including addressing equality of outcomes concerns in applications like credit scoring
models, which have faced criticism for perpetuating discrimination against
underrepresented communities, such as racial minorities. By employing
ZKML, banks can make their creditworthiness assessments fully
transparent and trustworthy, guaranteeing that the same model is used for all users.
Auditors can evaluate model decisions alongside the proof provided by
users disputing their scores, thereby eliminating discrimination rooted in users'
backgrounds.

The remainder of this paper demonstrates the viability of our
approach. In the background section, we introduce ``snarkGPT,'' a verifiable ZKML pipeline. In the evaluation section, we provide empirical results that underscore the effectiveness
of our approach, specifically showcasing its applicability to a
GPT2-size model.

\section{Background}
\subsection{zkSNARKs} \label{background:zksnarks}
Zero-Knowledge Succinct Non-Interactive Arguments of Knowledge (zkSNARKs) \cite{bitansky2017hunting} represent advanced cryptographic tools that allow verifying the correctness of a computation without revealing the inputs or intermediate steps. The foundation of zkSNARKs is rooted in a rich mathematical framework characterised by their advantageous properties, including zero-knowledge, succinctness, non-interactivity, argument soundness, knowledge-soundness, and completeness.

In the field of verifiable computing (VC), zkSNARKs offer an effective solution by enabling clients to delegate computationally intensive tasks to providers. After delegation, the provider (prover) sends a concise cryptographic proof to the client (verifier), allowing the client to verify the correctness of the computation without the need for a full re-execution\cite{whyAndHowZKSNARKS}. Most notably, the size of the proof is significantly smaller than the size of the original computation. This succinct proof size is a fundamental characteristic of zkSNARKs, making them suitable for situations where executing the whole computation is not feasible, as is the case for generative AI. Moreover, in scenarios involving outsourced AI computations, zkSNARKs provide a robust mechanism for clients to verify the correctness of model inferences without compromising data confidentiality. This is particularly relevant in the context of secure model inference-as-a-service.

To describe zkSNARKs for generative AI, we can define them formally as, given a function evaluation $f(x; w) = y$, where $x$ is a public input, $w$ is a private input called \textit{witness} and $y$ the output, zkSNARKs allow the prover to generate a proof $\pi$ such that the verifier that knows $x$, $y$ and $\pi$ can check that the prover knows $w$ such that $f(x; w) = y$ \cite{https://doi.org/10.48550/arxiv.2210.08674}.
For example, in the case of transformers, $x$ can represent an input prompt, $w$ the model weights and $y$ the output of executing the model using input prompt $x$ and weights $w$. \\

Most zkSNARK protocols proceed in three steps \cite{nitulescu2020zk}:
\begin{enumerate}
    \item Arithmetisation: producing a system of polynomial equations over a large prime field (an arithmetic circuit) for which finding a solution is equivalent to computing  $f(x; w)$. Therefore, given $(f, y, x, w)$, the circuit constraints are met if and only if $y = f(x; w)$.
    \item Building an information-theoretic proof system, that is, building a proof system that guarantees soundness even against a computationally unbounded prover. This system usually relies on idealized components (oracles) or is inefficient.
    \item Compiling the information-theoretic proof system into an \textit{efficient} one using cryptographic tools at the cost of considering only computationally bounded adversaries.
\end{enumerate}
There are two main classes of zkSNARKs: Quadratic Arithmetic Program (QAP)-based and Polynomial Interactive Oracle Proof (PIOP)-based. QAP-based zkSNARKs utilize arithmetic circuits and divisibility checks. PIOP-based zkSNARKs compile circuits into polynomial constraints and employ polynomial commitment schemes. Recent years have seen a dominance of PIOP-based SNARKs mainly due to their increased flexibility.

This work uses one of the most popular and well-documented PIOP-based proof systems: \textit{Halo2}.

\subsubsection{Halo2}
Halo2 \cite{halo2Book} is an instance of a \textit{PIOP} made non-interactive via the Fiat-Shamir heuristic \cite{fiatShamir}. The scheme uses Plonkish arithmetisation \cite{gabizon2019plonk} to represent a relation and a satisfying witness to that relation as a low-degree polynomial. Then, the constructed polynomial is fed into a polynomial commitment scheme where the prover commits to the polynomial and can evaluate it provably at arbitrary points from the verifier. Halo2 uses the inner product argument \cite{innerProduct} as a polynomial commitment scheme. However, other frameworks using Halo2 can choose to have a different commitment scheme, as is the case of EZKL (introduced subsequently in the background section) using KZG \cite{KZG}. Using Halo2, a verifier can use an accumulation scheme to batch instances it wants to evaluate. 

Plonkish arithmetisation allows polynomial constraints with certain restricted forms of randomness and supports custom gates and lookup arguments. Plonkish arithmetisation conceptualises the arithmetic circuit as a matrix (referred to as \textit{circuit matrix}) of m columns and n rows over a given finite field $\mathbb{F}$ (therefore, the cells contain elements of $\mathbb{F}$). Each column \textit{j} encodes a wire and corresponds to a Lagrange interpolation polynomial $p_{j}(X)$ over the powers (rows) of an $n^{th}$ primitive of unity $\omega$ that evaluates to $p_{j}(\omega^{i}) = x_{ij}$. A permutation argument is used to enforce the equality of cells.

The matrix defines a sequence of polynomial constraints, which are
multivariate polynomials over $\mathbb{F}$ that must evaluate to zero for each row. In a polynomial constraint, the variables can refer to a cell in a given column of the current row or another row relative to this one.

To improve performance, Halo2 can trade memory for CPU by pre-computing and storing lookup tables for some part of the computation. The lookup argument enforces a relation between variables, where the relation is expressed as a table. The lookup table consists of two advice columns and two fixed columns of the matrix, where every expression in the set of advice columns is equal to some expression in the set of fixed columns. Therefore, the {lookup argument} is a more permissive version of the permutation argument.
It enforces that all the input-output pairs in the witness are valid input-output pairs in the fixed columns. Lookup tables are often leveraged to represent more complex arithmetic relationships in the circuit matrix that would otherwise be hard to capture only with additions and multiplications. A prime example is non-linearities in neural networks that would otherwise have to be numerically approximated.

\subsubsection{EZKL} \label{ezkl}

EZKL \cite{ezkl, ezklDocs} is a Rust library and command-line tool that allows constructing zkSNARKs for the inference phase of machine-learning models.
EZKL uses Halo2 with KZG in the backend, with certain modifications that make the library compatible with larger models, compared to existing solutions implementing zkSNARKs for machine learning models.
EZKL converts pre-trained models from ONNX to Halo2 circuit matrices. Instead of
directly implementing each of the 100+ ONNX operations individually, EZKL builds upon the tract
library \cite{tractSonos}, which reduces and
prunes complex ONNX graphs to compositions of a smaller set of
operations. By leveraging the Einstein summation operation, EZKL can
represent numerous linear operations such as matrix multiplication, dot
product, transposition, and tensor contraction with a single operation.
This approach allows EZKL to support a wide range of models, including
those with LSTM and self-attention layers while only having to support 20 operations as equivalent circuit constraints. Notably, EZKL improves the memory costs of the Halo2 circuit matrix by allocating additional columns if all of the available rows have been filled. By the definition of Halo2, there can be arbitrarily many columns in the circuit matrix, but the number of rows has to be a power of 2. This comes at the cost of increasing the proving time.

\subsection{Generative AI: from GPT down to nanoGPT}

Text generation within generative AI relies on advanced models such as Transformers, introduced in 2017 \cite{gpt1}, which have since become foundational for language models. The architecture consists of 12 transformer blocks with masked multi-head self-attention. GPT-2 and GPT-3 are improved versions with larger datasets and more parameters. 

GPT-2 largely follows the architecture of GPT-1 and has four trained models available for use, each having a different number of blocks and embedding sizes. Their number of parameters are 117M, 345M, 762M and 1.5B, respectively. GPT-3 \cite{gpt2} follows the same architecture as GPT-2, with a few hyperparameters modified to improve performance. The model has configurations ranging from 175M to 175B parameters. GPT-4 extends this line of models and accepts multiple modalities, but its authors have yet to document its architecture thoroughly in a peer-reviewed publication.

In this work, we focus on the GPT-2 architecture (the same as the GPT-3 architecture), as it is the latest architecture in the GPT series for which a Python implementation has been made public at the time of writing. We propose a zkSNARK for the transformer architecture as a proof-of-concept, using nanoGPT: a smaller version of the GPT-2 architecture. NanoGPT's code aims to be shorter and easier to interpret \cite{nanoGPT, minGPT}, but remains compatible with the original GPT-2 and can be used to train or finetune medium-sized GPTs.

\section{Related work}

\subsection{ZKML} \label{subsection:relatedWorkZKML}
Previous research has proposed zkSNARK protocols or the inference phase of smaller neural networks.
Lee et al. \cite{lee2020vcnn} propose a protocol for verifiable convolutional neural networks, using QAP-based zkSNARKs for pooling and ReLU, QPP (Quadratic Polynomial Program)-based zkSNARKs for convolutions and CP(Commit-and-Prove)-SNARKs for interconnecting the layers. They argue that their scheme improves the key generation/proving time by 25X compared to the state-of-the-art zkSNARK scheme \cite{cryptoeprint:2016/260} for a small example of the MNIST model consisting of a single convolutional layer with ReLU and a single pooling layer. Moreover, for VGG16, they argue that their scheme improves the performance by 1800X, compared with  \cite{cryptoeprint:2016/260}.

Another approach was proposed by Weng et al. \cite{DBLP:journals/corr/abs-2201-09186}, for privacy-preserving and verifiable CNNs by using a modified version of QAP - QMP (Quadratic Matrix Program). The authors use a QMP-based arithmetic circuit to express convolutional relations and generate zkSNARKs proofs based on Homomorphic Encryption (HE) and collaborative inference, which promises to protect both the model and the data privacy. They argue that they obtain 17.6X faster \textit{Setup} time and 13.9X faster proving time than the QAP-based method in their experiments.

ZEN is an optimizing compiler for verifiable neural networks using zkSNARKs, introduced by Feng et al. \cite{feng2021zen}. It proposes two privacy-preserving, verifiable inference schemes for neural networks, $ZEN_{acc}$ and $ZEN_{infer}$, which promise to provide privacy guarantees for both the model and the data. The authors also introduce two optimizations for R1CS: \textit{R1CS friendly quantization}, which they argue "brings up to 73.9$\times$ savings in R1CS constraints for a convolution layer and up to 8.4$\times$ reduction for fully connected kernel without any additional accuracy loss" (\cite{feng2021zen}, page 2), and \textit{Stranded encoding of R1CS Constraints}, which they argue "leads to up to 2.2$\times$ improvement in R1CS constraints for convolution kernel and 3.4$\times$ improvement for fully connected kernel" (\cite{feng2021zen}, page 2).

Kang et al. \cite{https://doi.org/10.48550/arxiv.2210.08674} followed the steps of EZKL and also constructed Halo2 proofs for verifying the inference phase of a machine learning model. They propose an ImageNet-scale \cite{imagenet} zkSNARK using Halo2 and present three protocols for verifying the ML model accuracy, verifying the ML model predictions and trustless retrieval of items matching a predicate, respectively. They argue that they can achieve up to 79\% accuracy on ImageNet-scale while simultaneously taking as few as 10s and 5,952 bytes to verify and that the zkSNARK can be scaled down to take as few as 0.7s to verify
at 59\% accuracy. Comparing their results to \cite{lee2020vcnn}, \cite{DBLP:journals/corr/abs-2201-09186}, \cite{feng2021zen} and \cite{zkCNN}, they claim that the proving time for the prior work is at least
10X higher than their Halo2 method and up to 1,000X higher.

\subsection{Fairness}

Fairness has garnered significant attention within the field of machine
learning, encompassing a wide range of investigations into the
unintended behaviours exhibited by machine learning models
\cite{barocas2017fairness}. The literature has extensively explored
various facets of fairness, including the concepts of group fairness and
performance fairness. The former strives to mitigate model bias with
respect to specific protected attributes
\cite{du2021fairnessaware, rodriguez2021enforcing, chu2021fedfair},
while the latter necessitates the uniformity of performance
distributions across different recipients
\cite{zhang2020fairfl, deng2020distributionally, pentyala2022privfairfl}.
This section concentrates on the notion of performance fairness.

Performance fairness has garnered particular prominence within Federated Learning, wherein client data originating from multiple
sources exhibits high heterogeneity and spans geographical diversity.
Mohri et al.~introduced an initial framework for optimizing the
performance of the poorest-performing device through a minimax
optimization scheme \cite{mohri2019agnostic}. Subsequently, the q-FedAvg
algorithm was introduced by Li et al.~\cite{li2020fair}, offering a more
flexible optimization objective tailored to achieve varying
degrees of fairness. More recently, Ditto was introduced as an approach
to engender fairness by enabling the learning of personalized models
\cite{li2021ditto}.

As we can see, the works mentioned above primarily centre on narrowing
the performance gap across heterogeneous clients. This pursuit aligns
with our approach, which is complementary. Our method empowers clients to identify and conscientiously report
performance fairness disparities without compromising the sensitivity
and confidentiality of their private data.

\section{A provably fair GPT prototype: SNARKGPT} \label{section:snarkGPT}

This section describes ZKML in the context of performance fairness and how zkSNARKs can ensure that all clients of a generative AI service are treated equally. We demonstrate the practical viability of this solution by introducing \textit{snarkGPT}, a verifiable ZKML pipeline for the GPT-2 model, and suggest an accompanying protocol to capture potential fairness deviations.

\subsection{A protocol fostering performance fairness}

In the context of remote service hosting generative AI models, we set the challenge for a provider to instil confidence in consumers regarding the uniform quality and execution of the models it serves.

Starting from a simple case, we question how a premium user (or institution) of ChatGPT can obtain the guarantee that they use the premium version of the model at all times and not a cheaper version sometimes. With the growing popularity of ChatGPT, it is unclear how the service deals with the high number of requests. At the moment of writing, the only option for paying users is to trust that OpenAI always serves the premium model to premium users. 



 One option is for the model provider to share the model weights. However, this poses problems. Sharing these weights means leaking substantial intellectual property encapsulated within these models, and not all users may have the necessary computing power to use these weights effectively \cite{Butler_2023}.

We believe that a promising solution to the lack of guarantees in these cases is using zkSNARKs, which can prove, beyond a reasonable doubt, the correct execution of a computation. Indeed, as discussed in the related work section, previous work has proposed zkSNARKs solutions for verifying the correct execution and accuracy of other machine learning models, such as CNNs, DNNs and ImageNet-scale.

One possible ZKML protocol is as follows:
\begin{enumerate}
    \item The provider of a generative AI model publishes a commitment (for example, a hash) to the model weights using a Credible Commitment Device \cite{kalai2010commitment} (for instance in a public domain).
    \item A client can then send some input data to the provider and ask the provider to evaluate the model on that input data.
    \item The provider generates the output of the model evaluated on the client's data, as usual, and generates a zkSNARK. The proof takes as parameters the weights of the model (private), the client's input data (public) and the output of the model (public). The zkSNARK enables anyone to validate the correct execution of the model with significantly smaller costs. The proof attests that the model produces the reported output given the parameters above and the public commitment to the model.
    \item The client can then run a verification protocol to accept or reject the proof without access to the model weights.
    
\end{enumerate}

To evaluate the viability of ZKML to equip generative AI with performance fairness and quality guarantees, we introduce "snarkGPT", a zkSNARK prototype for the GPT-2 architecture in the following sections.

\subsection{nanoGPT adaptations} 
\begin{center}
\begin{table}[tb]
\caption{The architecture of the nanoGPT model \cite{nanoGPT}.}
\begin{tabular}{p{1.5cm}p{1.5cm}p{4cm}}
 \toprule
 \textbf{Layer} & \textbf{Parameters} & \textbf{Role} \\
 \hline
 Token \break Embedding   & vocabulary size,\break embedding size  & Creates token embeddings for the input sequence. \\
 \hline
 Positional \break Embedding &  block size,\break embedding size  &  Creates positional embeddings for the input sequence. \\
 \hline
 Dropout & dropout rate & Takes as input the token and positional embeddings and zeroes \textit{rate}\% random elements. \\
 \hline
 List\break[Transformer Blocks]    & number of layers & The model's hidden layers/transformer blocks. Takes as input the token and positional embeddings after dropout. \\
 \hline
 Layer Normalisation &   embedding size  & Applies Layer Normalization to the output of the hidden layers, as described in \cite{layerNormalization}. \\
 \hline
 Language Modelling Head (Linear Layer) & embedding size,\break vocabulary size  & A linear layer with weights tied to the input embedding. \\
 \bottomrule
\end{tabular}
\label{nanoGPTArchitecture}
\end{table}
\end{center}


The architecture of the nanoGPT model is illustrated in table \ref{nanoGPTArchitecture}. The model takes in as input an input prompt, which can be empty, and generates text based on the input sequence. Anchoring this setting to a medical diagnosis, a patient's symptoms represent the input prompt, and the medical prescription represents the generated text. To achieve quality guarantees, we must adapt the nanoGPT model to make it compatible with zkSNARKs.

First, since zkSNARKs operate over finite fields, all model values must be mapped to a value in the chosen finite field. In the self-attention mechanism of transformers, the elements in the upper-triangular portion of the self-attention matrix are set to $-\inf$ to eliminate the information that follows in the sequence. $-\inf$ is not a valid element of a finite field. Therefore, we set the elements in the upper-triangular portion of the self-attention matrix to the smallest value covered in the Halo2 lookup table. More specifically, let $B$ be the lookup table's logarithmic size. The lookup table stores all values in the given finite field between $-2^{B-1}+1$ and $2^{B-1}-1$. Thus, we zero out the elements in the self-attention matrix by setting them to $-2^{B-1}+1$. Then, we remove all optimizations that are not compatible with ONNX (flash attention \cite{flashAttention}) nor EZKL (array slicing, cube operation). Note that these operations are only used in the nanoGPT python code to accelerate computing self-attention on GPU. Removing them has thus no impact on the accuracy of the model.

\section{Empirical feasibility evaluation} \label{section:experiments}

\subsection{Overview}

This part of the study aims to shed light on the feasibility of our cryptographic solution for complementing fairness in machine learning models, specifically focusing on nanoGPT. As the time and memory costs for the verifier are constant in Halo2, we evaluate the performance of our proposed method based on the time and memory costs for the prover, who has to generate the zkSNARK proof. The server hosting the model would bear the burden of generating the proof when the user wants to obtain quality guarantees for the service they're consuming. Our experiments provide empirical evidence of the overhead and challenges in adopting ZKML as a de facto approach to improve performance fairness in a client-server setting.

We first test to what extent we can scale up two of the main components of our nanoGPT architecture when generating the proof: the number of layers and the size of the embeddings. We consider these parameters in our tests because they are core parameters of the self-attention mechanism. To be able to focus on the parameters we select to increase, we choose relatively small values for the rest of the parameters for simplicity. More specifically, we set the vocabulary size to 65, the block size to 64, the number of heads to 4, the batch size to 1 and the dropout rate to 0. All our tests are performed using a fixed version of the EZKL framework (commit \textit{8f122bf}\footnote{\url{https://github.com/zkonduit/ezkl/commit/8f122bf2eb5794b681364eb182d7cdd8a7350fe4}}). We perform all our experiments on a machine with the following specifications: Intel i7-9700K CPU, 3.60 GHz / 4.90 GHz, 64 GB 3600 MHz RAM and 200 GB Swap area in NVMe SSD.

We are also interested in how the circuit size (the log number of rows pre-allocated for the Halo2 circuit matrix to represent the model) influences the time and memory costs for the prover. Therefore, in our tests, we create multiple proofs for the same nanoGPT model configuration using EZKL, wherein we systematically vary the logarithmic dimensions allocated to the matrix representation of the circuit. We expect that a larger matrix will result in significantly higher proving costs.

\subsection{Scaling Up the Model}

 Tables \ref{table:scaleUpEmbeddings} and \ref{table:scaleUpLayers} present a detailed breakdown of the runtime incurred during proof generation for varying embedding sizes and layers, respectively. 
  As shown in Tab. \ref{table:scaleUpEmbeddings}, the computational demands appear to increase nonlinearly with the size of the embeddings. For instance, the proof generation time increases from $\sim$5 minutes for an embedding size of 64 to $\sim$20 minutes for an embedding size of 144. Similarly, an augmentation in the number of layers corresponds to an increase in the time required for proof generation. These combined results emphasize the complex factors that must be considered when optimizing the architecture of generative AI models within cryptographic frameworks. ZKML practitioners must balance the model's representational capacity with a tractable computational efficiency. In other words, the findings highlight the need for a nuanced approach that considers both the sophistication of the model's representation and the practical constraints imposed by zkSNARKs.

\subsection{Circuit Matrix Size Impact on Proof Generation}

Table \ref{table:MemoryImpact} illustrates the impact of the (logarithmic) size of the circuit matrix representation of the nanoGPT model on proof generation time and memory costs. That is, the number of rows we have to pre-allocate in memory for the circuit matrix representation when using EZKL and Halo2. As illustrated, the runtime exhibits a non-linear progression, initially improving from 3 minutes and 55 seconds to 3 minutes and 11 seconds as the logarithmic size increases from 14 to 18. However, beyond this point, there is a notable rise in runtime, with a substantial increase observed for a matrix logarithmic size of 24 and 25. The increased runtime for larger circuit matrices aligns with our expectations. Suppose we allocate significantly more rows in the matrix than needed for the number of polynomial constraints. In that case, those rows will be filled with 0s and still be used in the process (as each column describes a wire in the circuit), adding overhead.

Similarly,  an upward trend in memory cost is apparent as the logarithmic size of the circuit matrix increases. This is expected, as the entire matrix representation of the circuit is generated and stored in RAM when generating the proof. Notably, the memory cost experiences a more significant escalation than the runtime, reaching 148 GB for a matrix logarithmic size of 25. This finding underscores the intricate considerations involved in managing memory resources when dealing with larger circuit matrices.



\subsection{Generalisation beyond GPT architectures}\label{empirical-evaluation-of-constraint-generation}

In this section, we present the empirical results of our investigation
into the relationship between a model's architecture, the number of parameters and the number of
constraints generated within zkSNARK proofs. The number of constraints directly reflects the proving costs. It is thus crucial in our context to understand how generalizable our results are to other models beyond generative LLMs.

\subsubsection{Experimental Setup}\label{experimental-setup}

To assess the generality of the parameter-constraint relationship, we
designed experiments based on a model configuration previously examined
in Modulus Labs \cite{modulusLabs}. This configuration consists of
multiple Multi-Layer Perceptron (MLP) layers, each comprising a linear
layer, the Rectified Linear Unit (RELU) activation function, and a
scaling-down factor. We selected this setup to assess if the observed
parameter-constraint correlation extends to transformer models like
nanoGPT.

\subsubsection{Results}\label{results}

Our empirical findings reveal a striking contrast in constraint
generation between nanoGPT and the Modulus Labs MLP model. Specifically,
while the number of constraints (M) in the zkSNARK proof for the Modulus
Labs model roughly aligns with the number of parameters (N) in the
model, our nanoGPT model exhibits a significantly higher M / N ratio,
ranging from approximately 58X to 85X more constraints than parameters.
These observations are summarized in Table
\ref{tab:constraints_comparison}.



Notably, transformers such as nanoGPT exhibit a pronounced M / N ratio increase. For instance, with an embedding size of 64, the M / N
ratio approximates 64. In practical terms, a one-million-parameter model
generates a staggering 64 million constraints, while a 250,000-parameter
model still generates 16 million constraints. In stark contrast, a
four-layer convolutional network with 3,047 parameters generates merely
13,152 constraints, yielding an M / N ratio of approximately 4.

\subsubsection{Factors Impacting Constraint
Generation}\label{factors-impacting-constraint-generation}

We recognize that the time and memory costs associated with zkSNARK
proof generation depend not only on the number of parameters but also on
various architectural and methodological factors. Specifically, the M /
N ratio within a ZK-circuit is influenced by 1) the specific gates and
constraints chosen to represent neural network layers; 2) the network
architecture, as we have seen from the difference between transformers
and other types of networks; 3) the chosen proof system and its embedded
features, such as lookup arguments that may yield fewer constraints than
those without.

Addressing the challenge of minimizing this M / N ratio in zkSNARK proof
generation remains an open area of research and exploration. Our
findings underscore the need for innovative architectural designs and
proof methodologies to enhance the efficiency and practicality of our
cryptographic solution for fairness in machine learning models.

\section{Conclusion}

This work introduced a pioneering approach that leverages Zero-Knowledge Machine Learning (ZKML) to address the critical issues of performance fairness and reliability in generative AI models. By harnessing zkSNARKs, our protocol empowers users to confidently engage with AI models through remote cloud APIs, ensuring that the model they interact with aligns precisely with their expectations. 

The practical viability of our proposal is demonstrated through "snarkGPT," a verifiable ZKML pipeline for GPT2-like models. Our empirical evaluations indicate the effectiveness of our approach, particularly in scaling up the model and its generalization beyond GPT architectures. We showed how snarkGPT could be inserted in an end-to-end protocol where the user could obtain a proof of correct execution alongside the model's regular output and verify it in constant time regardless of the input size or model complexity.

This paper lays the foundation for a new era of transparent and reliable AI, emphasizing ZKML as a vital component in guaranteeing quality and achieving performance fairness. By leveraging ZKML, we bridge the gap between AI model providers and users, ensuring that AI is a truly equitable and transparent tool for all.

\begin{table*}[ht] 
\caption{Runtime during proof generation for nanoGPT configured with different embedding sizes.}
\centering
\begin{tabular}{p{3cm}cccccc}
\toprule
embeddings size & 64      & 80      & 96      & 112     & 128    & 144     \\ \hline
Runtime for proof generation & 5m 2s & 6m 7s & 8m 33s & 11m 46s & 16m 9s & 20m 28s\\
\bottomrule
\label{table:scaleUpEmbeddings}
\end{tabular}
\end{table*}

\begin{table*}[ht] 
\caption{Runtime during proof generation for nanoGPT configured with different layer configurations.}
\centering
\begin{tabular}{llllll}
\toprule
\multicolumn{1}{c}{number of layers} & \multicolumn{1}{c}{4} & \multicolumn{1}{c}{8} & \multicolumn{1}{c}{12} & \multicolumn{1}{c}{16} & \multicolumn{1}{c}{20} \\ \hline
Runtime for proof generation & 5m 2s               & 8m 30s               & 12m 19s                & 17m 4s                    & 29m 9s       \\
 \bottomrule
 \label{table:scaleUpLayers}
\end{tabular}
\end{table*}

\begin{table*}[ht]
\caption{Time and memory costs incurred during proof generation for nanoGPT for different sizes of the circuit matrix.}
\centering
\begin{tabular}{p{3cm}lllllll}
\toprule
log size of the circuit matrix & 14     & 16     & 18     & 20     & 22     & 24     & 25      \\ 
\hline
Runtime for proof generation & 3m 55s & 3m 26s & 3m 11s & 3m 34s & 4m 40s & 9m 53s & 14m 26s \\ 
\hline
Memory cost for proof generation & 44 GB  & 45 GB  & 46 GB  & 49 GB  & 63 GB  & 118 GB & 148 GB \\
\bottomrule
\label{table:MemoryImpact}
\end{tabular}
\end{table*}

\begin{table*}[ht]
\caption{The results of generating a zkSNARK proof for our nanoGPT model versus the Modulus Labs model.}
\centering
\begin{tabular}{ccccc}
\toprule
        & Number of parameters & Runtime    & Memory cost & Number of Constraints \\ \hline
        & 0.2 M                & 5m 2s        & 63 GB       & 17 M                  \\
        & 0.31 M               & 6m 7s        & 76 GB       & 24 M                  \\
nanoGPT & 0.45 M               & 8m 33s       & 108 GB      & 34 M                  \\
        & 0.79 M               & 11m 46s      & 132 GB      & 45 M                  \\
        & 1.01 M               & 16m 9s       & 173 GB      & 58 M                  \\ \hline
        & 0.2 M                &              &             & 0.2 M                 \\
        & 0.31 M               &              &             & 0.3 M                 \\
MLP     & 0.45 M               & $\sim$1m 30s & $\sim$26 GB & 0.45 M                \\
        & 0.79 M               &              &             & 0.8 M                 \\
        & 1.01 M               &              &             & 1 M \\
\bottomrule
\end{tabular}
\label{tab:constraints_comparison}
\end{table*}

\clearpage
\bibliography{references}

\end{document}